\documentclass[11pt,a4paper]{article}
\usepackage[hyperref]{acl2020}
\usepackage{times}
\usepackage{latexsym}
\usepackage{tabularx}
\usepackage{booktabs}
\usepackage{graphicx}

\usepackage{microtype}

\aclfinalcopy %
\def\aclpaperid{13} %

\title{MobIE: A German Dataset for Named Entity Recognition, Entity Linking and Relation Extraction in the Mobility Domain}

\author{Leonhard Hennig ~~~~~ Phuc Tran Truong ~~~~~
Aleksandra Gabryszak \\
\mbox{}\\
German Research Center for Artificial Intelligence (DFKI)\\
Speech and Language Technology Lab \\
\texttt{\{leonhard.hennig,phuc\_tran.truong,aleksandra.gabryszak\}@dfki.de}}

\date{}

\begin{document}
\maketitle
\begin{abstract}
We present \texttt{MobIE}, a German-language dataset, which is human-annotated with 20 coarse- and fine-grained entity types and entity linking information for geographically linkable entities. The dataset consists of 3,232 social media texts and traffic reports with 91K tokens, and contains 20.5K annotated entities, 13.1K of which are linked to a knowledge base. A subset of the dataset is human-annotated with seven mobility-related, n-ary relation types, while the remaining documents are annotated using a weakly-supervised labeling approach implemented with the Snorkel framework. To the best of our knowledge, this is the first German-language dataset that combines annotations for NER, EL and RE, and thus can be used for joint and multi-task learning of these fundamental information extraction tasks. We make \texttt{MobIE} public at \url{https://github.com/dfki-nlp/mobie}.

\end{abstract}

\section{Introduction}
Named entity recognition (NER), entity linking (EL) and relation extraction (RE) are fundamental tasks in information extraction, and a key component in numerous downstream applications, such as question answering~\cite{yu_improved_2017} and knowledge base population~\cite{ji_knowledge_2011}. Recent neural approaches based on pre-trained language models (e.g., BERT~\cite{devlin-etal-2019-bert}) have shown impressive results for these tasks when fine-tuned on supervised datasets \cite{akbik_contextual_2018,de_cao_autoregressive_2021,alt_improving_2019}. However, annotated datasets for fine-tuning information extraction models are still scarce, even in a comparatively well-resourced language such as German~\cite{benikova2014b}, and generally only contain annotations for a single task (e.g., for NER CoNLL'03 German~\cite{tjong2003}, GermEval 2014~\cite{benikova2014b}; entity linking GerNED~\cite{ploch-etal-2012-gerned}). In addition, research in multi-task~\cite{ruder_overview_2017} and joint learning~\cite{sui_joint_2020} has shown that models can benefit from exploiting training signals of related tasks. To the best of our knowledge, the work of~\citet{schiersch-2018} is the only dataset for German that includes two of the three tasks, namely NER and RE, in a single dataset.

\begin{figure*}[t!]
    \centering
    \includegraphics[width=\linewidth,clip,trim=0 12 0 0]{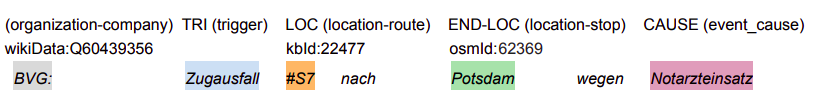}
    \caption{Traffic report annotated with entity types, entity linking and arguments of a \textit{Canceled Route} event}.
    \vspace{-0.3cm}
    \label{fig:example-traffic-report}
    
\end{figure*}

In this work, we present \texttt{MobIE}, a German-language information extraction dataset which has been fully annotated for NER, EL, and n-ary RE. The dataset is based upon a subset of documents provided by~\citet{schiersch-2018}, but focuses on the domain of mobility-related events, such as traffic obstructions and public transport issues. Figure~\ref{fig:example-traffic-report} displays an example traffic report with a \textit{Canceled Route} event. All relations in our dataset are n-ary, i.e.\ consist of two or more arguments, some of which are optional. Our work expands the dataset of~\citet{schiersch-2018} with the following contributions:
\begin{itemize}
    \setlength\itemsep{1pt}
    \item We significantly extend the dataset with 1,686 annotated documents, more than doubling its size from 1,546 to 3,232 documents
    \item We add entity linking annotations to geo-linkable entity types, with references to Open Street Map\footnote{\url{https://www.openstreetmap.org/}} identifiers, as well as geo-shapes
    \item We implement an automatic labeling approach using the Snorkel framework~\cite{ratner_snorkel_2017} to obtain additional high quality, but weakly-supervised relation annotations
\end{itemize}
The dataset setup allows for training and evaluating algorithms that aim for fine-grained typing of geo-locations, entity linking of these, as well as for n-ary relation extraction. The final dataset contains $20,484$ entity, $13,104$ linking, and $2,036$ relation annotations.

\begin{figure*}[t!]
    \centering
    \includegraphics[width=\textwidth,clip,trim=0 0 0 0]{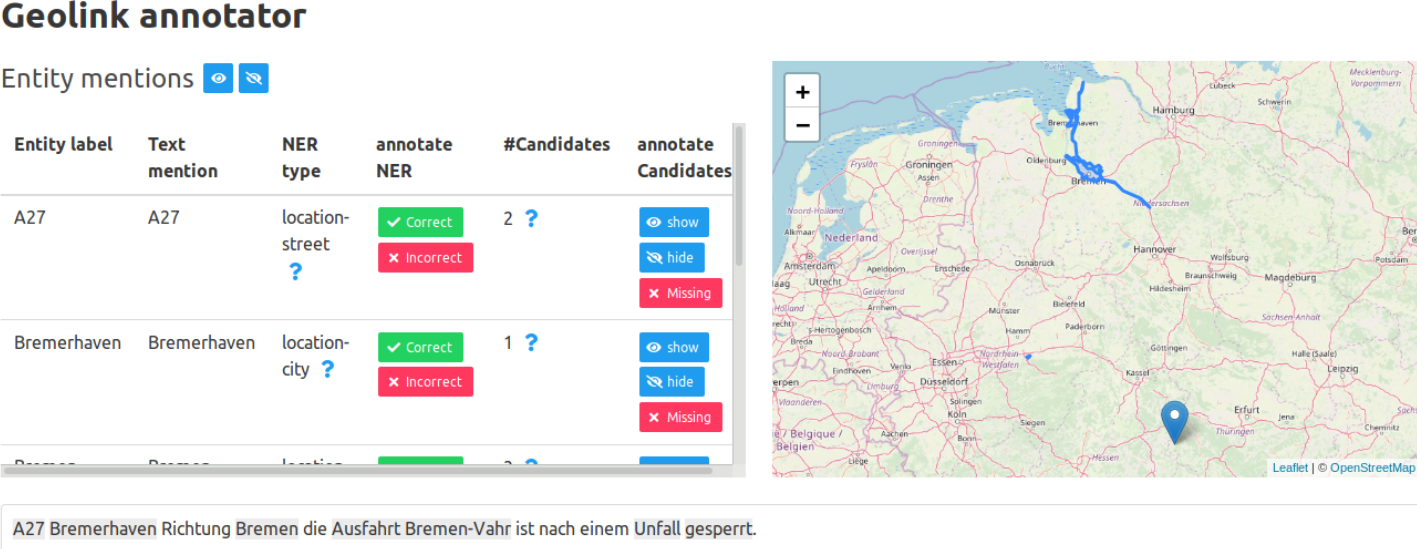}
    \caption{Geolinker: Annotation tool for entity linking}
    \label{fig:geolinker}
    \vspace{-0.3cm}
\end{figure*}

\section{Data Collection and Annotation}
\label{sec:annotation}

\subsection{Annotation Process}

We collected German Twitter messages and RSS feeds based on a set of predefined search keywords and channels (radio stations, police and public transport providers) continuously from  June 2015 to April 2019 using the crawlers and configurations provided by~\citet{schiersch-2018}, and randomly sampled documents from this set for annotation. The documents, including metadata, raw source texts, and annotations, are stored with a fixed document schema as AVRO\footnote{\url{avro.apache.org}} and JSONL files, but can be trivially converted to standard formats such as CONLL. Each document was labeled iteratively, first for named entities and concepts, then for entity linking information, and finally for relations. For all manual annotations, documents are first annotated by a single trained annotator, and then the annotations are validated by a second expert. All annotations are labeled with their source, which e.g.\ allows to distinguish manual from weakly supervised relation annotations (see Section~\ref{sec:events}).

\subsection{Entities}
\label{sec:entities}

Table~\ref{tab:entity_stats} lists entity types of the mobility domain that are annotated in our corpus. All entity types except for \textit{event\_cause} originate from the corpus of~\citet{schiersch-2018}. The main characteristics of the original annotation scheme are the usage of coarse- and fine-grained entity types (e.g., \textit{organization}, \textit{organization-company}, \textit{location}, \textit{location-street}), as well as trigger entities for phrases which indicate annotated relations, e.g., \textit{
``Stau'' (``traffic jam'')}. We introduce a minor change by adding a new entity type label \textit{event\_cause}, which serves as a label for concepts that do not explicitly trigger an event, but indicate its potential cause, e.g., \textit{``technische Störung'' (``technical problem'')} as a cause for a \textit{Delay} event.

\subsection{Entity Linking}
\label{sec:linking}

In contrast to the original corpus, our dataset includes entity linking information. We use Open Street Map (OSM) as our main knowledge base (KB), since many of the geo-entities, such as streets and public transport routes, are not listed in standard KBs like Wikidata. We link all geo-locatable entities, i.e.\ \textit{organizations} and \textit{locations}, to their KB identifiers, and external identifiers (Wikidata) where possible. We include geo-information as an additional source of ground truth whenever a location is not available in OSM\footnote{This is mainly the case for \textit{location-route} and \textit{location-stop} entities, which are derived from proprietary KBs of Deutsche Bahn and Rhein-Main-Verkehrsverbund. Standardized ids for these entity types, e.g.\ DLID/DHID, were not yet available at the time of creation of this dataset.}. Geo-information is provided as points and polygons in WKB format\footnote{\url{https://www.ogc.org/standards/sfa}}.

Figure~\ref{fig:geolinker} shows the annotation tool used for entity linking. The tool displays the document's text, lists all annotated geo-location entities along with their types, and a list of KB candidates retrieved. The annotator first checks the quality of the entity type annotation, and may label the entity as \textit{incorrect} if applicable. Then, for each valid entity the annotator either labels one of the candidates shown on the map as correct, or they select \textit{missing} if none of the candidates is correct.

\subsection{Relations}
\label{sec:events}
\begin{table}[t!]
\small
\centering
\begin{tabular}{ll}
\toprule
 \textbf{Relation} & \textbf{Arguments} \\
\midrule
\emph{Accident} &  \textsc{default-args}, delay \\\hline
\emph{Canceled Route} & \textsc{default-args} \\\hline
\emph{Canceled Stop} & \textsc{default-args}, route \\\hline
\emph{Delay} & \textsc{default-args}, delay \\\hline
\emph{Obstruction} &  \textsc{default-args}, delay \\\hline
\emph{Rail Repl. Serv.} &  \textsc{default-args}, delay \\\hline
\emph{Traffic Jam} & \textsc{default-args}, delay, jam-length \\\hline
\bottomrule
\end{tabular}
\caption{Relation definitions of the \texttt{MobIE} dataset. \textsc{default-args} for all relations are: location, trigger, direction, start-loc, end-loc, start-date, end-date, cause. Location and trigger are essential arguments for all relations, other arguments are optional.}
\label{tab:relations}
\end{table}
Table~\ref{tab:relations} lists relation types and their arguments. The relation set focuses on events that may negatively impact traffic flow, such as \textit{Traffic Jam}s and \textit{Accident}s. All relations have a set of required 
and optional arguments, and are labeled with their annotation source, i.e., human or weakly-supervised. Different relations may co-occur in a single sentence, e.g.\ \textit{Accident}s may cause \textit{Traffic Jam}s, which are often reported together.

\textbf{Human annotation}. The annotation in \citet{schiersch-2018} is performed manually. Annotators labeled only explicitly expressed relations where all arguments occurred within a single sentence. The authors report an inter-annotator agreement of $0.51$ (Cohen's $\kappa$) for relations.

\textbf{Automatic annotation with Snorkel.} To reduce the amount of  labor required for relation annotation, we explored an automatic, weakly supervised labeling approach. Our intuition is that due to the formulaic nature of texts in the traffic report domain, weak heuristics that exploit the combination of trigger key phrases and specific location types provide a good signal for relation labeling. For example, \textit{
``A2 Dortmund Richtung Hannover 2 km Stau''} is easily identified as a \textit{Traffic Jam} relation mention due to the occurrence of the \textit{``Stau''} trigger in combination with the road name \textit{``A2''}.

\begin{figure}[t!]
    \centering
    \includegraphics[width=\linewidth,clip,trim=0 0 0 0]{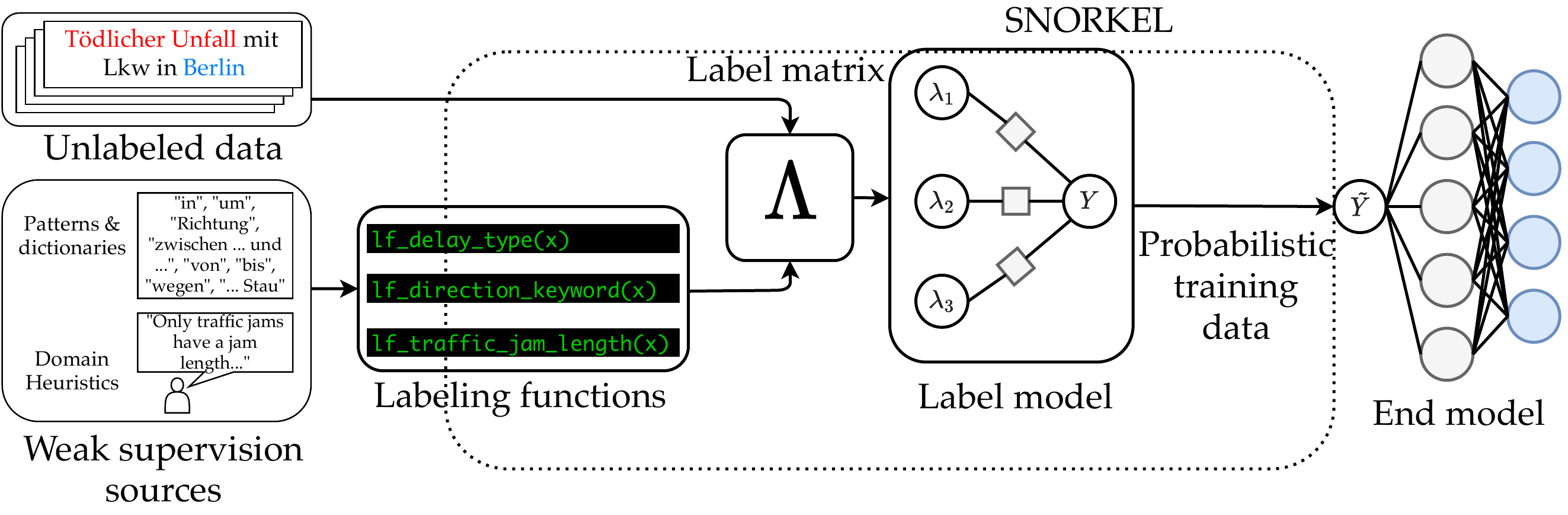}
    \caption{Snorkel applies user-defined, `weak' labeling functions (LF) to unlabeled data and learns a model to reweigh and combine the LFs’ outputs into probabilistic labels.}
    \label{fig:snorkel-workflow}
    \vspace{-0.3cm}
\end{figure}

We use the Snorkel weak labeling framework~\cite{ratner_snorkel_2017}. Snorkel unifies multiple weak supervision sources by modeling their correlations and dependencies, with the goal of reducing label noise~\cite{ratner_data_2016}. Weak supervision sources are expressed as labeling functions (LFs), and a label model combines the votes of all LFs weighted by their estimated accuracies  and outputs a set of probabilistic labels (see Figure~\ref{fig:snorkel-workflow}). 

We implement LFs for the relation classification of trigger concepts and role classification of trigger-argument concept pairs. The output  is used to reconstruct n-ary relation annotations. Trigger classification LFs include keyword list checks as well as examining contextual entity types. %
Argument role classification LFs are inspired by~\citet{chen-ji-2009-language}, and include distance heuristics, entity type of the argument, event type output of the trigger labeling functions, context words of the argument candidate, and relative position of the entity to trigger. We trained the Snorkel label model on all unlabeled documents in the dataset that contained at least a \textit{trigger} entity (690 documents). The probabilistic relation type and argument role labels were then combined into n-ary relation annotations. 

We verified the performance of the Snorkel model using a randomly selected development subset of 55 documents with human-annotated relations. On this dev set, Snorkel-assigned trigger class labels achieved a F1-score of $80.6$ (Accuracy: $93.0$), and role labeling of trigger-argument pairs had a F1-score of $72.6$ (Accuracy: $83.1$). This confirms our intuition that for the traffic report domain, weak labeling functions can provide useful supervision signals.

\section{Dataset Statistics}
\label{sec:corpus}
\begin{table}[t!]
\centering
\small
\begin{tabular}{lrrr}
\toprule
 & \textbf{Twitter} &    \textbf{RSS} &  \textbf{Total} \\
\midrule
\# docs      &   2,562 &    670 &  3,232 \\
\# sentences &   5,409 &  1,668 &  7,077 \\
\# tokens    &  62,330 & 28,641 & 90,971 \\
\# entities  &  13,573 &  6,911 & 20,484 \\
\# linked    &   8,715 &  4,389 & 13,104 \\
\# events    &   1,461 &    575 &  2,036 \\
\bottomrule
\end{tabular}
\caption{Dataset statistics per source}
\label{tab:datasets_stats}
\vspace{-0.3cm}
\end{table}

We report the statistics of the \texttt{MobIE} dataset in Table~\ref{tab:datasets_stats}.  %
The majority of documents originate from Twitter, but RSS messages are longer on average, and typically contain more annotations (e.g., $10.3$ entities/doc versus $5.3$ entities/doc for Twitter). The annotated corpus is provided with a standardized \emph{Train/Dev/Test} split. To ensure a high data quality for evaluating event extraction, we include only documents with manually annotated events in the \emph{Test} split. 

Table~\ref{tab:entity_stats} lists the distribution of entity annotations in the dataset, Table~\ref{tab:entity_linking_stats} the distribution of linked entities. Of the $20,484$ annotated entities covering 20 entity types, $13,104$ \textit{organization*} and \textit{location*} entities are linked, either to a KB reference id, or marked as NIL. The remaining entities are non-linkable types, such as time and date expressions. The fraction of NILs among linkable entities is $43.1$\% overall, but varies significantly with entity type. \textit{Locations} that could not be assigned to a specific subtype are more often resolved as NIL. A large fraction of these are highway exits (e.g. \textit{``Pforzheim-Ost''}) and non-German locations, which were not included in the subset of OSM integrated in our KB. In addition, candidate retrieval for \textit{organization}s often returned no viable candidates, especially for non-canonical name variants used in tweets.

The dataset contains $2,036$ annotated traffic events, $1,280$ manually annotated and $756$ obtained via weak supervision. Table~\ref{tab:event_stats} shows the distribution of relation types. 
\textit{Canceled Stop} and \textit{Rail Replacement Service} relations occur less frequently in our data than the other relation types, and \textit{Obstruction} is the most frequent class. 
\begin{table}[ht!]
\centering
\small
\begin{tabular}{lrrr}
\toprule
 & \textbf{Twitter} &   \textbf{RSS} & \textbf{Total} \\
\midrule
date                 &     434 &   549 &   983 \\
disaster-type        &      78 &    18 &    96 \\
distance             &      37 &   175 &   212 \\
duration             &     413 &   157 &   570 \\
event-cause          &     898 &   116 & 1,014 \\
location             &     887 & 1,074 & 1,961 \\
location-city        &     844 & 1,098 & 1,942 \\
location-route       &   2,298 &   324 & 2,622 \\
location-stop        &   1,913 & 1,114 & 3,027 \\
location-street      &     634 &   612 & 1,246 \\
money                &      16 &     3 &    19 \\
number               &     527 &   198 &   725 \\
org-position         &       4 &     0 &     4 \\
organization         &     296 &   121 &   417 \\
organization-company &   1,843 &    46 & 1,889 \\
percent              &       1 &     0 &     1 \\
person               &     135 &     0 &   135 \\
set                  &      18 &    37 &    55 \\
time                 &     683 &   410 & 1,093 \\
trigger              &   1,614 &   859 & 2,473 \\
\bottomrule
\end{tabular}
\caption{Distribution of entity annotations}
\label{tab:entity_stats}
\end{table}

\begin{table}[ht!]
\centering
\small
\begin{tabular}{lrrr}
\toprule
 & \textbf{\# entities} &  \textbf{\# KB} & \textbf{\# NIL} \\
\midrule
location             &      1,961 &   703 & 1,258 \\
location-city        &      1,942 & 1,486 &   456 \\
location-route       &      2,622 & 2,138 &   484 \\
location-stop        &      3,027 & 1,898 & 1,129 \\
location-street      &      1,246 & 1,036 &   210 \\
organization         &        417 &     0 &   417 \\
organization-company &      1,889 &   192 & 1,697 \\
\bottomrule
\end{tabular}
\caption{Distribution of entity linking annotations}
\label{tab:entity_linking_stats}
\end{table}

\begin{table}[ht!]
\centering
\small
\begin{tabular}{lrrr}
\toprule
     & \textbf{Twitter} & \textbf{RSS} & \textbf{Total} \\
\midrule
Accident               &     316 &  80 &   396 \\
Canceled Route          &     259 &  75 &   334 \\
Canceled Stop           &      25 &  42 &    67 \\
Delay                  &     337 &  48 &   385 \\
Obstruction            &     386 & 140 &   526 \\
Rail Replacement Service &      71 &  27 &    98 \\
Traffic Jam             &      67 & 163 &   230 \\
\bottomrule
\end{tabular}
\caption{Distribution of relation annotations}
\label{tab:event_stats}
\end{table}

\section{Conclusion}
We presented a dataset for named entity recognition, entity linking and relation extraction in German mobility-related social media texts and traffic reports. Although not as large as some popular task-specific German datasets, the dataset is, to the best of our knowledge, the first German-language dataset that combines annotations for NER, EL and RE, and thus can be used for joint and multi-task learning of these fundamental information extraction tasks. The dataset is freely available under a CC-BY 4.0 license at \url{https://github.com/dfki-nlp/mobie}.

\section*{Acknowledgments}
We would like to thank Elif Kara, Ursula Strohriegel and Tatjana Zeen for the annotation of the dataset. This work has been supported by the German Federal Ministry of Transport and Digital Infrastructure as part of the project DAYSTREAM (19F2031A), and by the German Federal Ministry of Education and Research as part of the project CORA4NLP (01IW20010).

\bibliography{references}
\bibliographystyle{acl_natbib}

\end{document}